\documentclass{article}

\PassOptionsToPackage{numbers, compress}{natbib}

\usepackage[final]{nips_2018}

\usepackage[utf8]{inputenc} 
\usepackage[T1]{fontenc}    
\usepackage{hyperref}       
\usepackage{url}            
\usepackage{booktabs}       
\usepackage{amsfonts}       
\usepackage{nicefrac}       
\usepackage{microtype}      

\usepackage{graphicx}

\usepackage{amsmath}
\DeclareMathOperator*{\argmax}{arg\,max}

\title{Annotation-cost Minimization for Medical Image Segmentation using Suggestive Mixed Supervision Fully Convolutional Networks}

\author{
      Yash S.~Bhalgat \thanks{These authors contributed equally to this work.}\\
  Department of Computer Science\\
  University of Michigan, Ann Arbor\\
  \texttt{yashsb@umich.edu} \\
  \And
  Meet P.~Shah \footnotemark[1] \\
  Department of Electrical Engineering\\
  Indian Institute of Technology - Bombay\\
  \texttt{meetshah1995@ee.iitb.ac.in} \\
  \And
  Suyash P.~Awate\\
  Department of Computer Science \& Engineering\\
  Indian Institute of Technology - Bombay\\
  \texttt{suyash@cse.iitb.ac.in} \\
}

\begin{document}

\maketitle

\begin{abstract}
    For medical image segmentation, most fully convolutional networks (FCNs) need strong supervision through a large sample of high-quality dense segmentations, which is taxing in terms of costs, time and logistics involved. This burden of annotation can be alleviated by exploiting weak inexpensive annotations such as bounding boxes and anatomical landmarks. However, it is very difficult to \textit{a priori} estimate the optimal balance between the number of annotations needed for each supervision type that leads to maximum performance with the least annotation cost.
    To optimize this cost-performance trade off, we present a budget-based cost-minimization framework in a mixed-supervision setting via dense segmentations, bounding boxes, and landmarks. We propose a linear programming (LP) formulation combined with uncertainty and similarity based ranking strategy to judiciously select samples to be annotated next for optimal performance. In the results section, we show that our proposed method achieves comparable performance to state-of-the-art approaches with significantly reduced cost of annotations.
\end{abstract}

\section{Introduction}

Image segmentation is ubiquitous in many medical image analysis applications. Unlike natural image analysis, obtaining high-quality per pixel annotations for medical images is relatively cumbersome due to the high costs, time and logistics involved. Previous works attempt to alleviate this annotation burden by either judiciously sampling the labelled datapoints in an active learning setting \cite{gu2014active} or by leveraging weak, inexpensive forms of annotations such as bounding boxes, anatomical landmarks to assist segmentation in a mixed-supervision setting \cite{shah2018ms}.

In active learning settings, a FCN starts with an initial set of images with corresponding annotations and iteratively suggests images in the remaining unlabelled dataset to be next annotated.
Annotations are requested on images for which the network is highly uncertain about it's predictions and which are dissimilar to the already annotated images until a predefined annotation budget is reached.

In mixed-supervision settings, a FCN performing dense segmentation is augmented with auxiliary tasks like object detection and anatomical landmark localization which need inexpensive annotations to assist the base task of segmentation. Multi-task networks in \cite{he2017mask, shah2018ms} have shown that adding complementary tasks enables the network learn better features and hence perform better on base task. 

Costs involved in annotating each form of supervision are significantly different and therefore knowing the optimal balance between the number of annotations needed for each supervision type is essential for cost-effective segmentation. For practical applications, it would be good to have a budget-constraint framework that enables suggestive annotation in mixed supervision settings. We propose a linear programming (LP) inspired cost-minimization framework to enable suggestive budget-constraint segmentation in mixed supervision settings.

\section{Methodology}
We have 3 modes of supervision in this particular application namely - dense segmentation (s), landmarks (l), and bounding box detections (d). We use the MS-Net architecture which is constructed using two types of components: (i) a base network for full-resolution feature extraction and (ii) 3 sub-network extensions - one for each of $\{s,l,d\}$. We modify each sub-network to include a concrete dropout layer \cite{gal2017concrete} to enable uncertainty estimation.

In the proposed method, every mode of supervision has a cost of annotation $c \in \{C_d, C_s, C_l\}$ and a value $v \in \{V_d, V_s, V_l\}$ which is an estimate of marginal gain in IoU by adding one sample of an annotation type. The training starts with a subset of the training data which has all three annotations. Let this set be denoted by $X_{init} = \{ (x_1, y_{1s}, y_{1l}, y_{1d}), (x_2, y_{2s}, y_{2l}, y_{2d}), \dots\}$, where $y_{is}, y_{il}, y_{id}$ denote the 3 annotations for the $i^{th}$ image. We train the network on $X_{init}$ using all 3 of its sub-network extensions and progressively suggest images in the remaining unlabelled set to be next annotated. We call this approach Suggestive Mixed Supervision Network (SMS-Net).

\textbf{Uncertainty and similarity estimation} \; After initial training, we obtain predictions on the remaining images in the training set. Using the concrete dropout layer during inference, for each image $x$ from the remaining set, we compute an estimate of uncertainty for the predictions across each supervision mode - $u_s(x), u_l(x)$ and $u_d(x)$. Let $h(.)$ denote the features from the $2^{nd}$ last layer of the full-resolution stream of the MS-Net. We also compute the similarity of $x$ with $X_{init}$ as $S(x,X_{init})=\underset{y\in X_{init}}{\max}\kappa(h(x),h(y))$, where $\kappa$ is a gaussian kernel similarity measure. 

\label{LP} \textbf{LP Formulation} The objective is to maximize the
value of annotation given a cost budget. We also want to select the samples with high prediction uncertainty and the ones which are diverse from the current training set. We define a vector $z(x)$, where $z_i(x)=1$ if we provide the $i^{th}$ type of annotation to image $x$. To select the images from the remaining set and the annotation types to be requested for each image, we solve the following modified $0$-$1$ integer LP problem:
\[ \argmax_{z(x),x\in X} \quad \sum_{x \in X}\underbrace{V^Tz(x)}_{\text{value gain}} +  \underbrace{\lambda_u u(x)^Tz(x)}_{\text{uncertainty}} - \underbrace{\lambda_s s(x)}_{\text{similarity}} \quad \text{subject to:} \; \sum_{x \in X} \underbrace{C^Tz(x)}_{cost} \leq b\]
where $V=(V_s, V_d, V_l)$, $C=(C_s, C_d, C_l)$, $u=(u_s, u_d, u_l)$ and $b$ is the allocated budget for this update. We used the \texttt{cvxpy} package for this optimization. We obtain the $z$ vector for all the images in the remaining set. Depending on the $1$'s in $z(x)$, we annotate the image $x$, add it to $X_{init}$ and retrain the network with this updated training set. As shown in the results section, this cost minimization formulation enables higher performance at significantly reduced annotation budgets.
In each LP update, we update the estimates of $V_i$ depending on the $\Delta$IoU obtained. Suppose $|Y_i^{t+1}|$ is the number of samples that were given the annotation $i$ in the $(t+1)^{th}$ LP update. Then,
\[ V_i(t+1) = \alpha_i V_i(t) + (1-\alpha_i)\beta_i \Delta IoU \quad i \in \{s,l,d\} \]
\[ \alpha_i = \frac{|Y_{i}^{t}|}{|Y_{i}^{t}|+|Y_i^{t+1}|}, \beta_i = \frac{V_i|Y_i^{t+1}|}{V_s|Y_s^{t+1}|+V_l|Y_l^{t+1}|+V_d|Y_d^{t+1}|} \]



\section{Results}
In our experiments, we use the JSRT database \cite{shiraishi2000development} which has 247 high-resolution ($2048\times 2048$) chest radiographs, each with expert segmentations, 166 landmark annotations and bounding boxes  \citep{van2006segmentation} covering 5 anatomical structures. We compare our proposed SMS-Net with MS-Net \cite{shah2018ms} and suggestive FCN \cite{yang2017suggestive}. We define the max budget as the cost required to provide all the images with all 3 types of annotations. 
We start by randomly selecting 20\% of the training samples as $X_{init}$. We apply the LP update as described in section \ref{LP} at $K$ regular intervals in the entire training process. In each LP update, we assign $(\frac{1}{K})^{th}$ of the total available budget. We set $K=10$ in all our experiments.

\begin{figure}
    \centering
    \begin{minipage}{0.42\textwidth}
        \centering
        \includegraphics[width=0.98\textwidth]{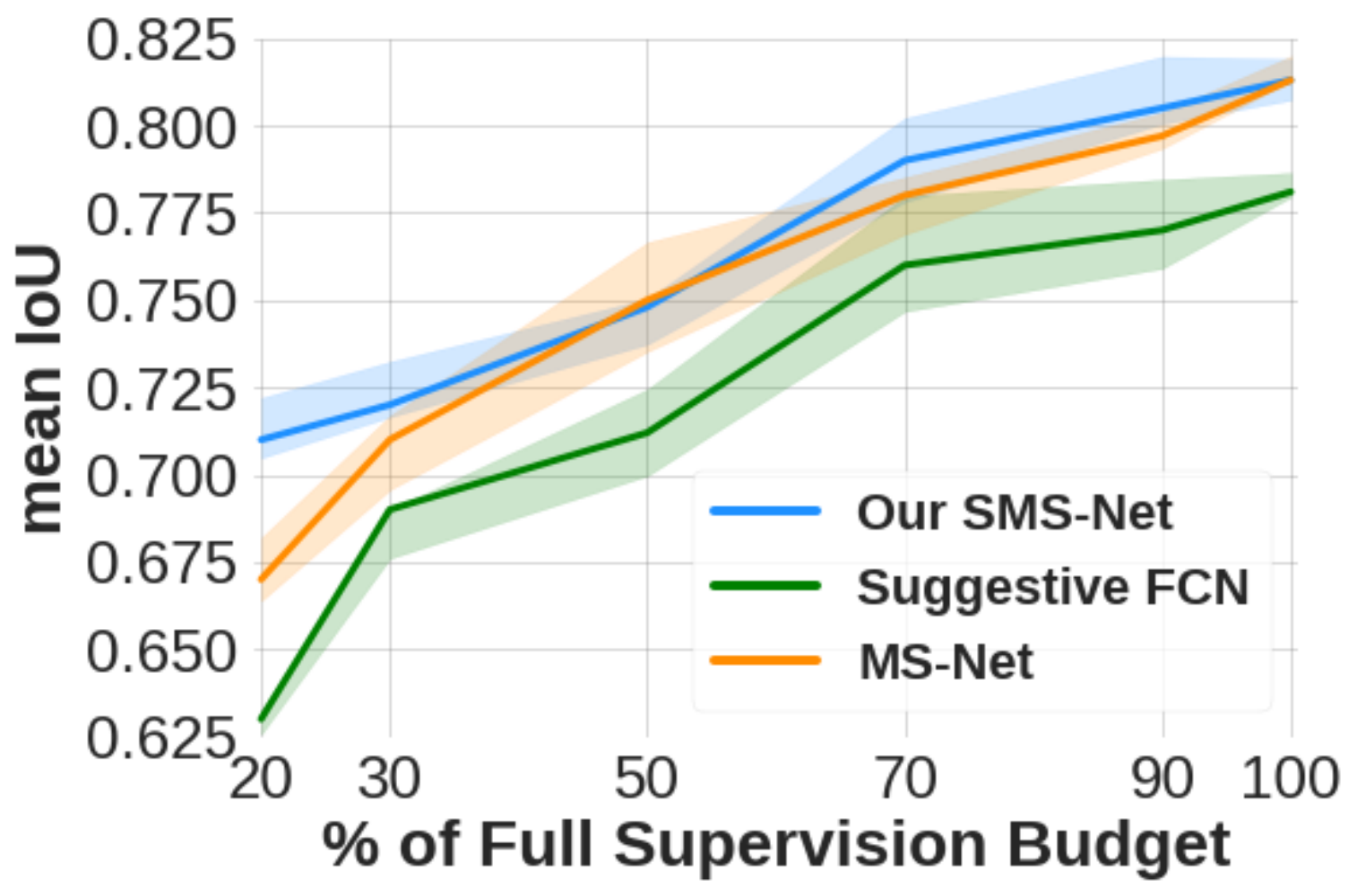}
        (a)

    \end{minipage}\hfill
    \begin{minipage}{0.4\textwidth}
        \centering
        \includegraphics[width=0.98\textwidth]{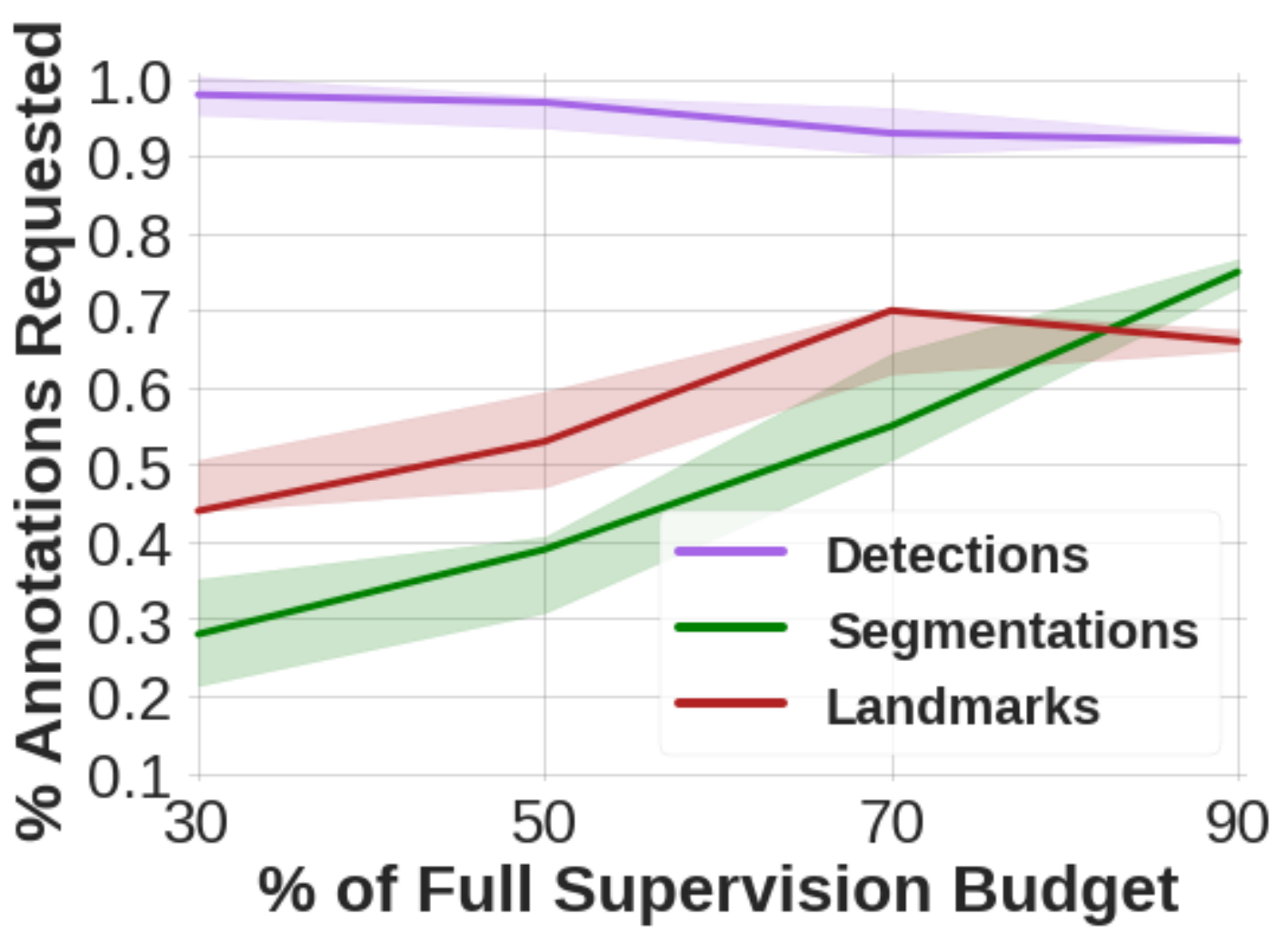}
        (b)
    \end{minipage}
    \caption{(a) Segmentation performance of different methods for varying levels of available budget (b) Percentage overall annotations requested by SMS-Net for varying levels of available budget. Full supervision budget is defined as the cost required to provide all the images with all 3 types of annotations}
    \label{fig:cost}
\end{figure}

In Figure \ref{fig:cost}(a), we present the mean IoU obtained for varying levels of available annotation budget. Since all annotations have different costs and values, 
Figure \ref{fig:cost}(b) shows the distribution of \textit{overall} samples drawn from each annotation type for varying levels of available budget. 
For all experiments, we set $C = (10, 1, 7)$. These costs are chosen to be proportionally in line with practical medical image annotation scenarios.
In Figure \ref{fig:cost}(b), we see that as we relax the budget constraint, more samples are provided with dense segmentations whereas lesser samples are given detections. At moderate and tighter budgets, detections and landmarks are utilized as they provide reasonable supervision at reduced costs. This selection strategy also leads to increase in performance as can be seen in Figure \ref{fig:cost}(a). 

\section{Conclusion and Future work}
We propose a method to enable suggestive annotation in mixed supervision settings for annotation cost-minimization in medical image segmentation. We show that our method achieves better performance compared to the state-of-the-art at significantly reduced annotation budgets. Future research directions would be to evaluate the proposed method on larger datasets with a non-fixed set of annotations and to build an end to end framework for joint optimization of the base segmentation architecture and cost minimization LP.

\bibliographystyle{plain}
\bibliography{nips_2018}


\end{document}